\documentclass[10pt,twocolumn,letterpaper]{article}

\usepackage{cvpr}
\usepackage{times}
\usepackage{epsfig}
\usepackage{graphicx}
\usepackage{amsmath}
\usepackage{amssymb}
\usepackage{multirow}

\usepackage[breaklinks=true,bookmarks=false]{hyperref}

\cvprfinalcopy % *** Uncomment this line for the final submission

\def\cvprPaperID{****} % *** Enter the CVPR Paper ID here

\begin{document}

%%%%%%%%% TITLE
\title{A Novel Approach to Artistic Textual Visualization via GAN}

\author{Yichi Ma\\
Saint Francis High School\\
Mountain View, CA, United States of America\\
{\tt\small yichima@sfhs.com}
\and
Muhan Ma\\
Beijing Normal University\\
Beijing, China\\
{\tt\small mamuhan.ma@gmail.com}
}

\maketitle
%\thispagestyle{empty}

%%%%%%%%% ABSTRACT
\begin{abstract}
	While the visualization of statistical data tends to a mature technology, the visualization of textual data is still in its infancy, especially for the artistic text. Due to the fact that visualization of artistic text is valuable and attractive in both art and information science, we attempt to realize this tentative idea in this article. We propose the Generative Adversarial Network based Artistic Textual Visualization (GAN-ATV) which can create paintings after analyzing the semantic content of existing poems.
	Our GAN-ATV consists of two main sections: natural language analysis section and visual information synthesis section. In natural language analysis section, we use Bag-of-Word (BoW) feature descriptors and a two-layer network to mine and analyze the high-level semantic information from poems. In visual information synthesis section, we design a cross-modal semantic understanding module and integrate it with Generative Adversarial Network (GAN) to create paintings, whose content are corresponding to the original poems. 
	Moreover, in order to train our GAN-ATV and verify its performance, we establish a cross-modal artistic dataset named "Cross-Art". In the Cross-Art dataset, there are six topics and each topic has their corresponding paintings and poems. The experimental results on Cross-Art dataset are shown in this article.
\end{abstract}

%%%%%%%%% BODY TEXT
\section{Introduction}
Data visualization \cite{liu2014survey} can create the visual representation of data, which plays an important role in analyzing and presenting data due to its intuitive expression. It can be categorized into statistical visualization \cite{kehrer2013visualization} and textual visualization \cite{reed2016generative,reed2016generating} based on the different modalities of data. As a branch of this discipline, statistical visualization can illustrate the statistical data via figures and there are many researchers concentrating on it with a number of approaches proposed. 

As an emerging branch, the research of textual visualization is still in the initial stage and there are a lot of challenges and problems in this field. The textual visualization refers to the technology, which is interested in translating text directly into image pixels or video frame pixels. For example, given a single-sentence human-written descriptions, the goal of textual visualization is to translate it to a synthetic picture whose content is related to the description. To realize this challenging textual visualization requires solving two problems: how to represent text for image synthesis and how to synthesize image via text representation. Specifically, we need to learn text representations which capture the important visual details and use these representations to synthesize compelling images which are as authentic as real images. Thanks to the strong capability of deep learning \cite{lecun2015deep} to learn representation, these problems about natural language representation and image synthesis have enormous progress in the past decades. 

However, there is a remaining issue that is not solved, namely the heterogeneity between text and image. The previous works employ the pairwise information provided by cross-modal datasets to learn the correlation between paired different modalities and then realize textual visualization. For example, in order to learn a textual visualization model, these methods require the training dataset to provide a real image as well as a description with the same visual content in an iteration. This is an effective practice, but needs a massive cost of labor to annotate the pairwise information, limiting the versatility of textual visualization.

Therefore, we want to propose an approach which is free to the numerous pairwise annotations and verify its ability in a specific application scenario. As is known to all, there is an interchangeability between the art forms with different modalities. For example, after reading a poem, artists can create a related painting. So it is often fantasized to create paintings from poems automatically via artificial intelligence system, which we define as “artistic textual visualization” (ATV). 

Nevertheless, to our best knowledge, there was no relevant research on ATV in the past and the existed textual visualization methods cannot be performed in this field because of the lack of pairwise annotations in art dataset. To solve this issue and realize ATV in this article. We propose the Generative Adversarial Network based Artistic Textual Visualization (GAN-ATV) which can create paintings after analyzing the semantic content of existing poems. This work is valuable and attractive to both art and information science.

\section{Methods}
\begin{figure*}
	\label{fig:overview}
	\begin{center}
		\includegraphics[width=0.9\linewidth]{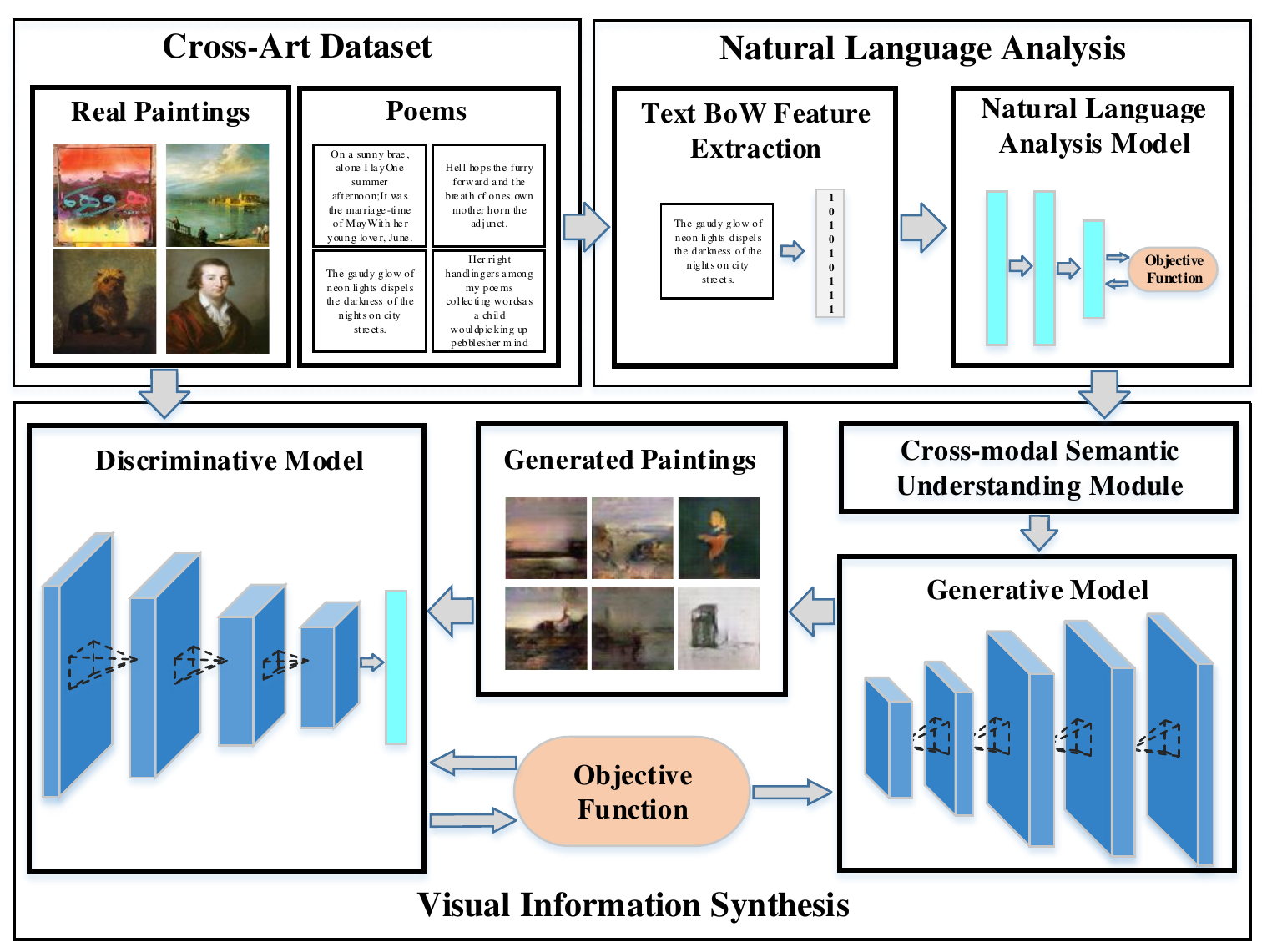}
	\end{center}
	\caption{The overview of our proposed GAN-ATV.}
\end{figure*}

The overview of our proposed GAN-ATV is presented in Figure \ref{fig:overview}. Our GAN-ATV receives text features of poems and real paintings as input and generate the created paintings. 

\begin{figure*}
	\begin{center}
		\includegraphics[width=0.9\linewidth]{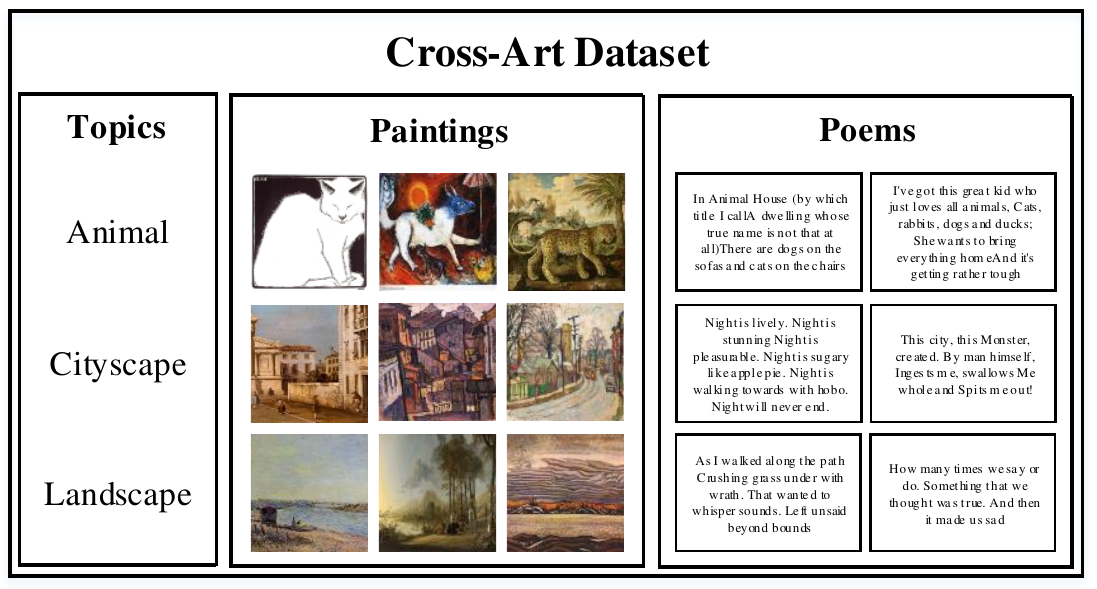}
	\end{center}
	\caption{Examples of Cross-Art Dataset.}
	\label{fig:dataset}
\end{figure*}

\subsection{Our Proposed GAN-ATV}
As we have mentioned in the “Introduction” section, textual visualization relies on the techniques of natural language representation and image synthesis and be promoted by deep learning. Our proposed GAN-ATV also builds on the deep learning and design a lot of artificial neural networks as the representation learning models. Our GAN-ATV consists of two main sections: natural language analysis and visual information synthesis. The natural language analysis section extracts the Bag-of-Words (BoW) feature descriptors \cite{harris1954distributional} from poems and has a three-layer network to further mine and analyze the high-level semantic information. The visual information synthesis section has a cross-modal semantic understanding module and integrates it with Generative Adversarial Network (GAN) \cite{goodfellow2014generative} to create paintings, whose content are corresponding to the original poems. 

\subsection{Natural Language Analysis}
The natural language analysis section is composed of text feature extraction and a natural language analysis model. We use the BoW feature descriptor as the original representations of poems. Given a set of poems, we firstly select the 2048 most frequent words in this set by statistics and associate them as a vocabulary. Then we regard a poem as a “bag”, which is a binary vector consisting of 0 or 1 in fact. If one most frequent words in this poem, the corresponding binary value will be set at 1 in the bag vector. Otherwise, the corresponding binary value will be set at 0. As we can see, the bag vector is a simplifying representation, disregarding grammar and even word order but keeping multiplicity. The natural language analysis model is a typical Artificial Neural Network (ANN) \cite{wang2003artificial} that is a computing system inspired by the biological neural networks that constitute animal brains. This kind of system can learn progressively from existing examples to realize tasks generally without task-specific programming. The main idea of ANN is usually called deep learning and is a nonlinear mapping from the original space to a specific space, such as semantic space for high-level semantic understanding. The natural language analysis model is a network with three hidden layers, which can mine and analyze the high-level semantic information from text. These three layers are fully-connected and the first two layers have 4096 hidden units, while hidden unit number of the third layer is same as number of poem topics. One hidden layer can be denoted as:

\begin{align}
y=tanh(wx+b)
\end{align}

where $y$ refers to the representation vector produced by this hidden layer, $x$ is the input vector, $w$ denotes the weight parameter of this hidden layer, and $b$ is the bias parameter. It should be noticed that $w$ is a matrix and the remaining variables are vectors. In order to learn the parameters $w$ and $b$, we design an objective function to learn the mapping from original feature descriptors to semantic representations and employ Stochastic Gradient Descent (SGD) \cite{bottou2010large} method to optimize this objective function. The objective function is implemented by minimizing the softmax loss function as follows:

\begin{align}
\min V(\theta)=\min(-\frac{1}{n}\sum_{i=1}^{n}\sum_{q=1}^{c}1\{L=q\}\log(\hat{p}(x,q,\theta)))
\end{align}

where $L$ is the label of instance $x$, $\theta$ refers to the parameters of network, and $c$ denotes the number of poem topics. If $L=q$, $1\{L=q\}$ equals to $1$, and otherwise $0$. $\hat{p}(x,q,\theta)$ is the probability distribution over topics and can be expanded as:

\begin{align}
\hat{p}(x,q,\theta)=\frac{e^{\theta_q\phi(x)}}{\sum_{l=1}^{c}e^{\theta_l\phi(x)}}
\end{align}

where $\phi(x)$ is the representation produced by the whole network. After optimizing this objective function by SGD, given a poem, natural language analysis section can mine and analyze its high-level semantic information. So we extract semantic information as probability vectors for the 1000 poems in testing set, representing probability that each poem corresponds to each topic.

\subsection{Visual Information Synthesis}
Visual information synthesis section can be divided into a cross-modal semantic understanding module and a visual generative adversarial module. The cross-modal semantic understanding module is not a neutral network but an interface. It receives probability vectors and generates noise vectors as “inspiration” with the same number of received probability vectors. Then it combines probability vectors with noise vectors one-by-one as the initial vectors for the visual generative adversarial module. These initial vectors not only maintain the semantic information of poems, but also bring inspiration to the visual generative adversarial module, which is similar to the artist's creative process.

The visual generative adversarial module has two neutral network model: a generative model G and a discriminative model D. The generative model captures the data distribution, and the discriminative model estimates the probability that a sample came from the training data rather than the generative model. Both the generative model and the discriminative model could be a nonlinear mapping function, such as a multi-layer perceptron. Specifically, the discriminative model is composed of four convolutional layers and one fully-connected hidden layer behind. The convolutional layer is a down-sampling mapping, which applies convolutional computation on the input matrix. After training, the higher level convolutional layer can produce a matrix representation with higher level semantic characteristics. So the last fully-connected hidden layer can easily extract semantic information from the last matrix representations and produce probability vectors representing whether the input data are real paintings or created paintings of a certain topic. The generative model takes the initial vectors produced by cross-modal semantic understanding module as input. And it consists of five deconvolutional layers. The deconvolutional layers are up-sampling mappings, which applies deconvolutional computation on the input matrix or vector. After training, the higher level deconvolutional layer can produce a matrix representation that is more similar to the real paintings.

As for the training of the visual generative adversarial module, we firstly pre-train the discriminative model and the generative model by original data and then train these models by an adversarial paradigm. At a training epoch, the generative model creates a batch of paintings from the initial vectors produced by cross-modal semantic understanding module, integrates them with real paintings, and then lets the discriminative model recognize whether these paintings are real to a certain topic. 

The model D and G play the following two-player minimax game with objective function:

\begin{equation}
	\begin{split}
		\min_{G}\max_{D}V(D,G)=& E_{x\sim p_{data}(x)}[\log(D(x|y))]\\
		& +E_{x\sim p_{z}(z)}[\log(1-D(z|y))]
	\end{split}
\end{equation}

In this equation, $E_{x\sim p_{data}(x)}[\log(D(x|y))]$ refers to the $D$ recognition result of real paintings, while $E_{x\sim p_{z}(z)}[\log(1-D(z|y))]$ is the recognition result of paintings created by $G$. The goal of generative model $G$ is to minimize this objective function, which means to fool the recognition process of discriminative model $D$. The aim of discriminative model $D$ is to maximize this objective function, which means to get the strong capability of predicting the authenticity and topic of painting data. They can boost each other by this adversarial training process.

By training our GAN-ATV, we can finally get an end-to-end model, which can convert the existed poems to the created paintings like artists’ creative process.

\section{Experiments}
The experiments of our proposed GAN-ATV consists of three main parts: the establishment of Cross-Art Dataset, the implementation and evaluation of natural language analysis, and the implementation and analysis of visual information synthesis. We use TensorFlow software library \cite{abadi2016tensorflow} and Python programming language to implement both natural language analysis and visual information synthesis sections. Moreover, at the end of experiments, we use our proposed GAN-ATV to create 1000 paintings for the 1000 poems in the testing set of Cross-Art dataset. We will analyze the 1000 created paintings as our final experimental results and show some created paintings.

\begin{figure*}
	\begin{center}
		\includegraphics[width=1\linewidth]{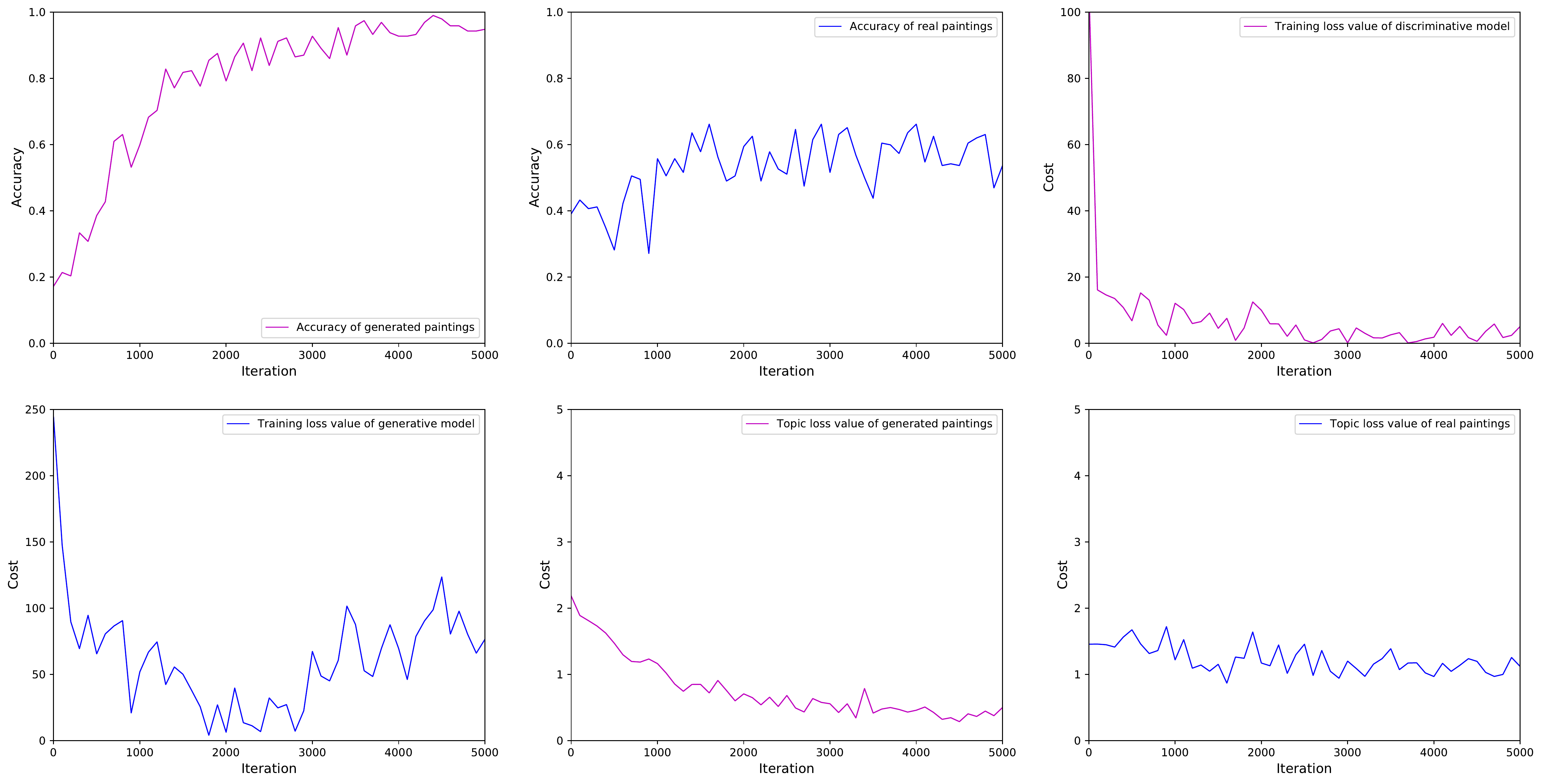}
	\end{center}
	\caption{Analysis of visual information synthesis.}
	\label{fig:experiment}
\end{figure*}

\subsection{Cross-Art Dataset}

In order to train our GAN-ATV and verify its performance, we establish a cross-modal artistic dataset named "Cross-Art". In the Cross-Art dataset, there are six topics and each topic has their corresponding paintings and poems. The six topics respectively are “abstract”, “animal”, “cityscape”, “landscape”, “human”, and “religion”. We design two web crawlers to collect poems and paintings respectively. The poems in Cross-Art dataset are crawled from PoemHunter.com, while the paintings are from WikiArt.org. There is no pairwise information between poems and paintings in Cross-Art dataset for two reasons. On one hand, it is almost impossible to make a poem for a painting artificially due to the artistic professionalism and high cost of labor. On the other hand, we want to verify the performance of our GAN-ATV in an application scenario without pairwise information annotations. Therefore, the poems and paintings do not have a one-to-one relationship and the number of poems can be different from the paintings. We collect 2814 poems and 61622 paintings for Cross-Art dataset and divide the 2814 poems into a training set with 1814 poems and a testing set with 1000 poems.

We find the fact that there is an interchangeability between the art forms with different modalities. The poems and paintings created by different artists have a lot of similar contents or styles, which is the interchangeability between the art forms with different modalities. For example, there is a poem describing a city’s night view via an anthropomorphic style and its title is City that Does Not Sleep. We can find a painting named Melody of the Night, depicting a city’s night view by an impressionist style, which can create a similar feeling for the viewer. This finding proves the innovation of our proposed approach reasonable in art.

In addition, we find that there is no existed poem have a natural paired painting, and vice versa. So nobody can establish a multimodal artistic dataset with natural pairwise information. This finding proves the necessity of our approach on the artistic textual visualization issue from the side.

\subsection{Implementation and Evaluation of Natural Language Analysis}
In the realization of natural language analysis, we find some phenomenon and improve our approach according to the phenomenon. Firstly, we find the ANN is a more effective method than other machine learning methods in text semantic content understanding. In the initial implementation, we apply the Support Vector Machine (SVM) \cite{cortes1995support} to realize natural language analysis model, which is a widely accepted generic model in data analysis. However, due to the high-level semantics in poems, the performance of SVM in predicting topics for poems is far worse than the ANN’s. In other word, if a poem describes an animal by an anthropomorphic style, the SVM may predict the topic of this poem as “human”, while the ANN can predict it as “animal”. This finding proves the rationality of our implementation.

Secondly, we evaluate the natural language analysis section quantitatively and verify its performance. We use the classification accuracy as the evaluation metric, which is a generic metric. The classification accuracy is a proportion of correctly predicted samples in all samples. The classification accuracy of the natural language analysis section for poems in testing set is 0.7130, which is a promising performance. This evaluation can illustrate that the natural language analysis section has a good understanding of the poems, which is a very critical implementation in artistic textual visualization.

\subsection{Implementation and Analysis of Visual Information Synthesis}
In the realization of visual information synthesis, we also find some interesting phenomenon and record some statistical data of intermediate results to analyze their trend in the training stage. We record six intermediate results at the 1 to 5000 iterations of model training process as follows: 

\subsubsection{Accuracy of generated paintings}
This is the accuracy of topics predicted by discriminative model for the generated paintings, which presents the semantic intensity of the paintings created by the generative model. This accuracy has been rising in the whole training process and is close to 1 at the end. That is to say the discriminative model holds the view that created paintings have a strong semantics in the end of the training stage.

\subsubsection{Accuracy of real paintings}
This is the accuracy of topics predicted by discriminative model for the real paintings, which indicates the ability of discriminative model to distinguish the topics of real paintings. This accuracy is rising but not stable due to the diversity of real paintings. 

\subsubsection{Topic loss value of generated paintings}
Topic loss value measures the training progress for the topic distinguishing ability of discriminative model and the lower value means more mature training progress. So the topic loss value of generated paintings presents the fitting degree of the discriminative model to the data distribution of generated paintings. This topic loss value has been decreasing below 1 at the end, which means the discriminative model can fit the data distribution of generated paintings well in the end of training process.

\subsubsection{Topic loss value of real paintings}
The topic loss value of real paintings expresses the fitting degree of the discriminative model to the data distribution of real paintings and it is also decreasing but not below 1. This result proves the negative influence of the diversity of real paintings to the discriminative model again.

\subsubsection{Training loss value of discriminative model}
Training loss value of discriminative model indicates the training progress of the authenticity distinguishing ability and the lower value means more mature training progress. This loss value is decreasing below 20 with great speed at the beginning and maintain below 10 in the end. It represents that the training process of discriminative model has a normal performance.

\subsubsection{Training loss value of generative model}
Training loss value of generative model indicates the training progress of the generated quality and the lower value is better. This loss value is also decreasing with great speed at the beginning but is not stable in the end, which is caused by the superior performance of the discriminative model on the created painting. This loss value will keep fluctuation, which is a manifestation of adversarial training process.

\subsection{The Created Paintings}
At the end of experiments, we use our proposed GAN-ATV to create 1000 paintings for the 1000 poems in the testing set of Cross-Art dataset. The 1000 created paintings are our final experimental results. We find that the most of created paintings have similar contents and styles with their corresponding source poems, which indicates our success in the realization of artistic textual visualization. 
\begin{figure*}
	\begin{center}
		\includegraphics[width=0.9\linewidth]{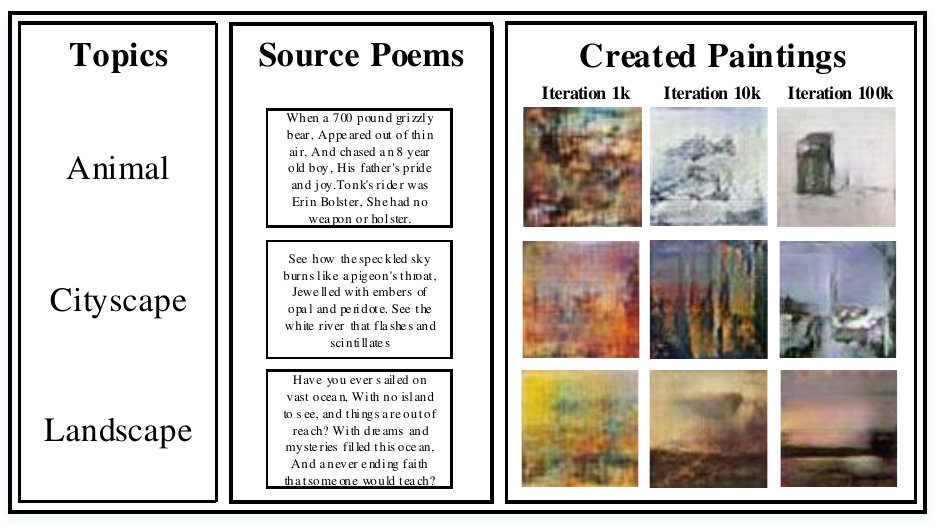}
	\end{center}
	\caption{Examples of created paintings at different iterations of training stage. We take the paintings at 100k iterations as final result.}
	\label{fig:generated_image}
\end{figure*}

To evaluate the final performance of our GAN-ATV, we propose a Semantic Correlation Factor (SCF), which can represent the correlation between original poems and created paintings. It can be calculated as follows:

\begin{align}
SCF=\frac{1}{n^2}(\sum_{i=1}^n1\{y(i)=p_o(i)\}\times\sum_{i=1}^n1\{y(i)=p_c(i)\})
\end{align}

where $p_o$ is correct semantic prediction of original poems and $p_c$ corresponds to created paintings. $n$ is the total number of original poems and $y$ denotes the semantic labels. Moreover, if $a=b$, $1{a=b}$ is equal to $1$, otherwise $0$. We present the SCF of GAN-ATV at different training iterations in Table 1. It can be noted that the performance of GAN-ATV is improved with the increasing of training iterations and it reaches the highest point at the end of training.

\begin{table}[htb]
	\begin{center}
		\begin{tabular}{|c|c|c|c|c|} 
			\hline
			Iterations & 100 & 1,000 & 10,000 & 100,000 \\
			\hline
			SCF & 0.3082 & 0.3974 & 0.4271 & 0.4865 \\
			\hline
		\end{tabular} 
	\end{center}
	\caption{SCF of GAN-ATV at different training iterations.}
	\label{table:SCF}
\end{table}

\section{Discussion}
As we can see in the “Experiment” section, our proposed GAN-ATV can use the training set of Cross-Art dataset to learning how to understand poems and create paintings for them. In this section, we will provide some interpretation of the results.

As is shown in Figure \ref{fig:experiment}, the adversarial models of visual information synthesis are trained with an expected trend. Therefore, we can get the 1000 created paintings for the 1000 poems in the testing set of Cross-Art dataset and show some examples in Figure \ref{fig:generated_image}. However, the accuracy of real paintings is rising but not stable, while it is a key metric to measure the ability of generative model. It is due to the fact that the real paintings in one same topic usually have lots of difference not only in content but also in style. For example, the “animal” topic has both beast and bird paintings and there are both realistic style and abstract style in this topic. It is noted that the creation of paintings is a really difficult task.

The Figure \ref{fig:generated_image} presents examples of created paintings at different iterations of training stage. We can see that the semantic contents of created paintings are clearer and clearer in the wake of iteration’s increasing. This is an expected phenomenon, which indicates the effectiveness of the visual information synthesis section. And we take the paintings at 100k iterations as final results, which is the end of training iterations.

However, there are some shortcoming in the final created paintings. Due to the limitation of Graphics Processing Unit (GPU) memory, our GAN-ATV can only create paintings in 64×64 resolution, which restricts the performance of GAN-ATV. We have attempted to solve this issue by super resolution technology \cite{dong2016image} which can replenish details of the created paintings and increase the resolution. But there are many problems in our implementation of super resolution and these problems are not solved in our experiment due to the deficiency of time. We will accomplish super resolution section as our future work.

To our best knowledge, our proposed GAN-ATV is the first ATV approach, because there was no relevant research on ATV in the past and the existed textual visualization methods cannot be performed in this field because of the lack of pairwise annotations in art dataset. Moreover, due to the fact that our GAN-ATV is free to the pairwise annotations in dataset, GAN-ATV is easy to extended to more application scenarios of textual visualization. We will also add this work into our future work to verify the versatility of our proposed GAN-ATV.

\section{Conclusions}
In this paper, we attempt to realize the tentative idea of artistic textual visualization and propose the Generative Adversarial Network based Artistic Textual Visualization (GAN-ATV). Our proposed approach can create paintings after analyzing the semantic content of existing poems. Moreover, in order to train our GAN-ATV and verify its performance, we establish a cross-modal artistic dataset named "Cross-Art". Then we conduct adequate experiments and analyze experimental results on Cross-Art dataset.

During this whole process, the most important thing I learned is how to carry out a scientific research project. Moreover, I have understood the theory of deep learning and adversarial learning, which not only lay the foundation for my future research life but also give me inspiration.
The future works lie in two aspects. On one hand, we will extend our approach to more application scenarios of textual visualization to verify its versatility. On the other hand, we attempt to improve the quality and resolution of created paintings via super resolution technology.

\clearpage
{\small
\bibliographystyle{ieee}
\bibliography{egbib}
}

\end{document}